\documentclass[times,twocolumn,final]{elsarticle}
\usepackage{cag}
\usepackage{framed,multirow}
\usepackage{caption}
\usepackage{subcaption}
\usepackage{amssymb}
\usepackage{latexsym}
\usepackage{url}
\usepackage{xcolor}
\definecolor{newcolor}{rgb}{0,0,1}
\usepackage{makecell}%set table fomat
\usepackage{hyperref}
\usepackage{array}
\usepackage[switch,pagewise]{lineno} %Required by command \linenumbers below
\usepackage{threeparttable}
\journal{Computers \& Graphics}
\RequirePackage{lineno}
\begin{document}
%\setpagewiselinenumbers
%\modulolinenumbers[1]
%\linenumbers
%\verso{Given-name Surname \textit{etal}}

\begin{frontmatter}
\graphicspath{{images/}}%set images path
\title{3D Shape Segmentation via Shape Fully Convolutional Networks\tnoteref{tnote1}}%

\author[1]{Pengyu Wang\corref{cor1}}
\cortext[cor1]{Equal contribution.}
\author[1]{Yuan Gan\corref{cor1}}
\author[1]{Panpan Shui}
\author[1]{Fenggen Yu}
\author[1]{Yan Zhang}
\author[2]{Songle Chen}
%% Third author's email
\author[1]{Zhengxing Sun}

\address[1]{State Key Lab for Novel Software Technology, Nanjing University}
\address[2]{Key Lab of Broadband Wireless Communication and Sensor Network Technology of Ministry of Education, Nanjing University of Posts and Telecommunications}
%%% Third author's email
%\ead{R.Laman@elsevier.com}
%\author[2]{Gail \snm{Rodney}}
%
%\address[1]{TU Wien, Favoritenstrasse 9-11/186, 1040 Wien, Austria}
%\address[2]{INESC-ID, Rua Alves Redol, 9 1000-021 Lisboa, Portugal}
%
%\received{1 February 2017}
%\finalform{28 March 2017}
%\accepted{2 April 2017}
%\availableonline{15 May 2017}
%\communicated{S. Sarkar}

\begin{abstract}
%%%
We design a novel fully convolutional network architecture for shapes, denoted by \emph{Shape Fully Convolutional Networks (SFCN)}. 3D shapes are represented as graph structures in the SFCN architecture, based on novel \emph{graph convolution and pooling operations}, which are similar to convolution and pooling operations used on images. Meanwhile, to build our SFCN architecture in the original image segmentation fully convolutional network (FCN) architecture, we also design and implement a \emph{generating operation} with bridging function. This ensures that the convolution and pooling operation we have designed can be successfully applied in the original FCN architecture. In this paper, we also present a new shape segmentation approach based on SFCN. Furthermore, we allow more general and challenging input, such as \emph{mixed datasets of different categories of shapes} which can prove the ability of our generalisation. In our approach, SFCNs are trained triangles-to-triangles by using three low-level geometric features as input. Finally, the feature voting-based multi-label graph cuts is adopted to optimise the segmentation results obtained by SFCN prediction. The experiment results show that our method can effectively learn and predict mixed shape datasets of either similar or different characteristics, and achieve excellent segmentation results.
%%%%
\end{abstract}

\end{frontmatter}

%\linenumbers
\section{Introduction}%1

%Begin By ShuiPP
Shape segmentation aims to divide a 3D shape into meaningful parts, and to reveal its internal structure. This is the basis and prerequisite to explore the inherent characteristics of the shape. The results obtained from shape segmentation can be applied to various fields of computer graphics, such as shape editing \cite{Yu:2004:MEP}, deformation \cite{Yang:2013:BAM}, and modelling \cite{Chen2015GMW}. Shape segmentation, therefore, has become a research hotspot, yet difficulties in the fields of digital geometric model processing and instance modelling persist.

Convolutional networks have shown excellent performance in various image processing problems, such as image classification \cite{Krizhevsky2012INC,Szegedy2015GDW,Simonyan2014VDC}, and semantic segmentation \cite{Cirean2012DNN,Farabet2013LHF,Pinheiro2013}. With the emerging encouraging study results, many researchers have devoted their efforts to various deformation studies on CNNs, one of which is the fully convolutional network (FCN) \cite{Long2015}. This method can train end-to-end, pixels-to-pixels on semantic segmentation, with no requirement over the size of the input image. Thus, it has become one of the key research topics in CNN networks.
 
\begin{figure*}[htbp]
  \centering
  \includegraphics[width=0.9\linewidth]{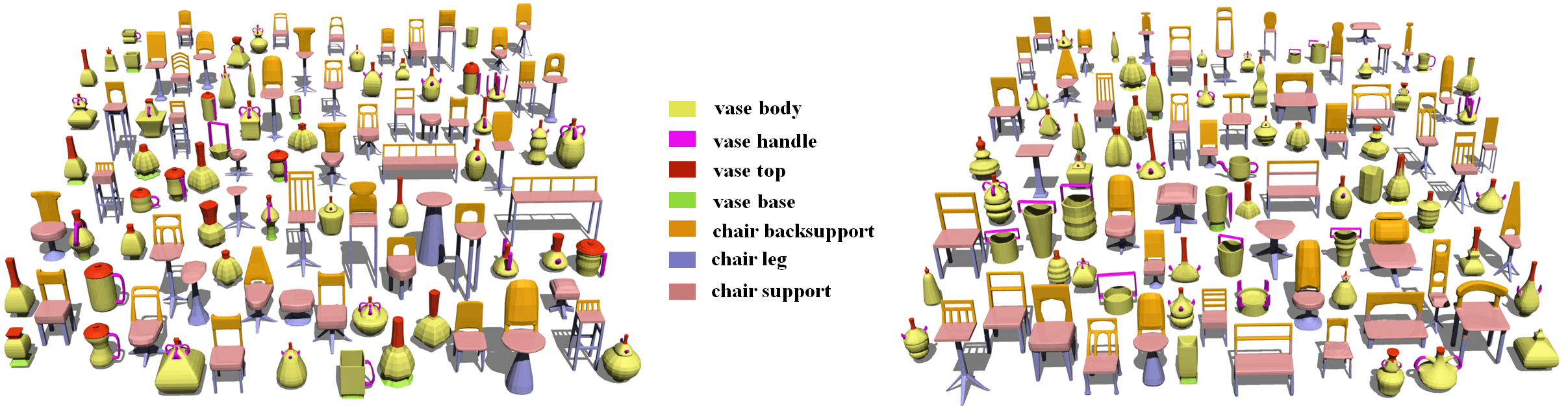}
  \caption{\label{fig:Fig0} Example of our shape segmentation results on one mixed shape dataset. The shapes on the left are part of the training set, and some segmentation results are shown on the right.}
\end{figure*}
%Figure \ref{fig:Fig1}% figure citation
\vspace{-5pt}

Although FCNs can generate good results in image segmentation, we cannot directly apply it to 3D shape segmentation. This is mainly because image is a type of a static 2D array, which has a very standard data structure and regular neighbourhood relations. Therefore, convolution and pooling can be easily operated when processing FCNs. However, the data structure of a 3D shape is irregular, so it cannot be directly represented as the data structure of an image. As triangle meshes have no regular neighbourhood relations like image pixels, direct convolution and pooling operations on a 3D shape is difficult to fulfil. 

\begin{figure}[htbp]
  \centering
  \includegraphics[width=0.9\linewidth]{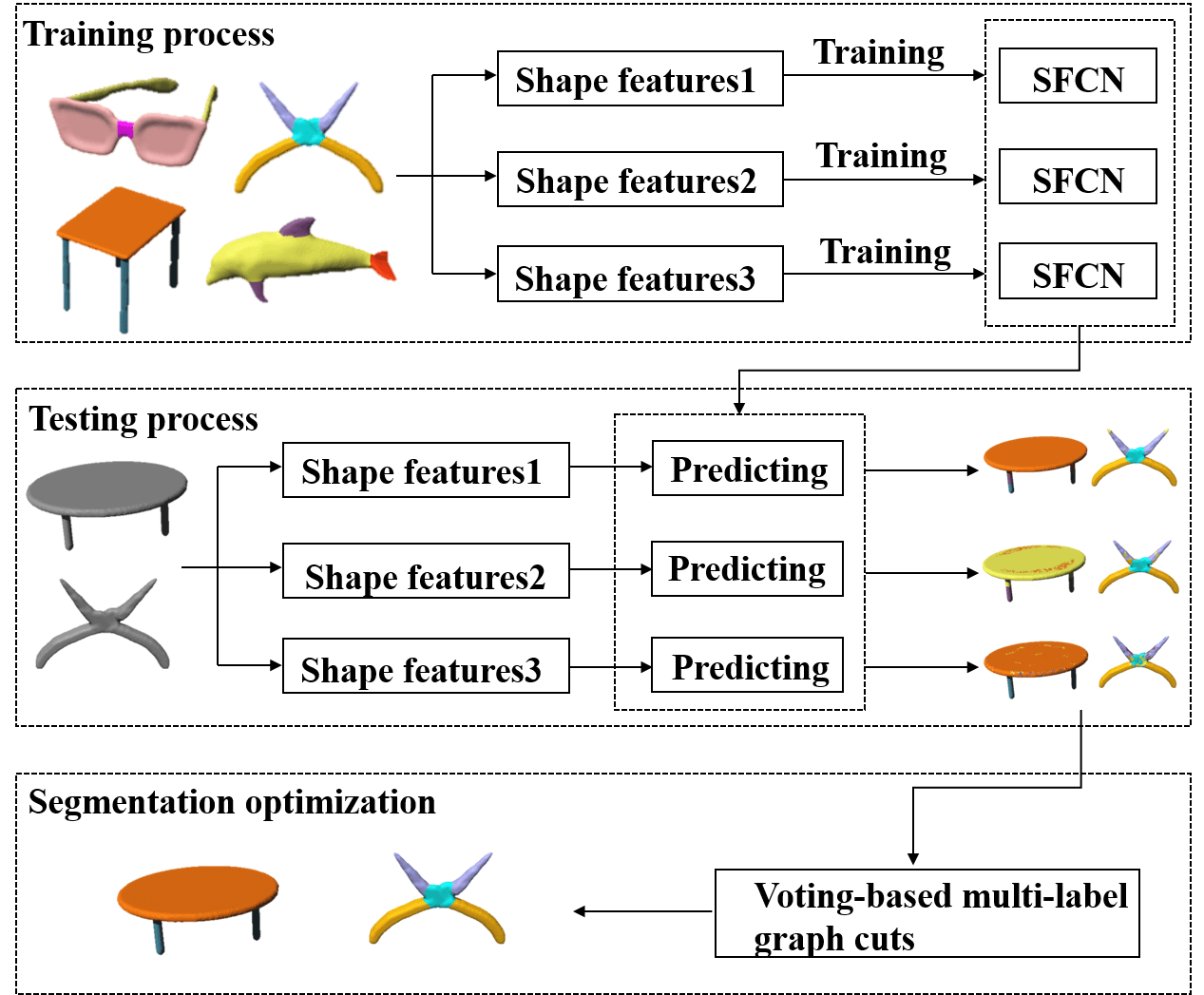}
  \caption{The pipeline of our method. It may be divided into 3 stages: training process, using the SFCN architecture to train under three different features; testing process, predicting the test sets through the SFCN architecture; optimisation process, optimising the segmentation results by the voting-based multi-label graph cuts method to obtain the final segmentation results.}\label{fig:Fig1}
\end{figure}

Inspired by the FCN architecture in image segmentation, we design and implement a new FCN architecture that operates directly on 3D shapes by coverting a 3D shape into a graph structure. Based on the FCN process of convolution and pooling operation on the image and existing methods of Graph Convolutional Neural Networks \cite{Edwards2016GBC,Niepert2016LCN,Defferrard2016CNN}, we design a shape convolution and pooling operation, which can be applied directly to the 3D shape. Combined with the original FCN architecture, we build a new shape fully convolutional network architecture and name it \emph{Shape Fully Convolutional Network (SFCN)}. Secondly, following the SFCN architecture mentioned above and the basic flow of image segmentation of FCN \cite{Long2015}, we devise a novel trained triangles-to-triangles model for shape segmentation. Thirdly, for higher accuracy of segmentation, we use the multiple features of the shape to complete the training on the SFCN. Utilising the complementarity between features, and combined with the multi-label graph cuts method \cite{Boykov2001,Kolmogorov2004WEF,Boykov2004AEC}, we optimise the segmentation results obtained by the SFCN prediction, through which the final shape segmentation results are obtained. Our approach can realize the triangles-to-triangles learning and prediction with no requirements on the triangle numbers of the input shape. Furthermore, many experiments show that our segmentation results perform better than those of existing methods \cite{Kalogerakis2010LTM,Guo:2015:TML,Xie2014TSS}, especially when dealing with a large dataset. Finally, our method permits mixed dataset learning and prediction. Datasets of different categories are combined together in the test, and the accuracy of the segmentation results of different shapes decreases very little. As shown in Figure \ref{fig:Fig0}, for a mixed shape dataset from COSEG \cite{Wang2012ACS} with several categories of shapes, part of the training set are displayed on the left, and some corresponding segmentation results are shown on the right.  Figure \ref{fig:Fig1} shows the process of our method.

%Figure \ref{fig:Fig1}% figure citation
\vspace{-5pt}

The main contributions of this paper include:
\begin{enumerate}
\setcounter{enumi}{0}
\item We design a fully convolutional network architecture for shapes, named Shape Fully Convolutional Network (SFCN), and is able to achieve effective convolution and pooling operations on the 3D shapes.
\item We present the shape segmentation and labelling based on SFCN. It can be triangles-to-triangles by three low-level geometric features, and outperforms the state-of-the-art shape segmentation.
\item Excellent segmentation results on training and predicting mixed datasets of different categories of shapes are achieved.
\end{enumerate}

\section{Related Work}%2

%Begin By ShuiPP
\textbf{Fully convolutional networks.} The fully convolutional network \cite{Long2015} proposed in 2015 is pioneering research which can effectively solve the problem of semantic image segmentation by pixel-level classification. Later, a great deal of research has emerged based on the FCN algorithms and achieved good results in various fields such as edge detection \cite{Xie2015HNE}, depth regression \cite{Liu2015LDS}, optical flow \cite{Dosovitskiy2015FLO}, simplifying sketches \cite{Simo2016LSF} and so on. However, the existing research on FCN is mainly restricted to image processing, largely because image has a standard data structure, easy for convolution, pooling and other operations. 

\textbf{Graphs CNN.} Standard CNN cannot work directly on data which have graph structures. However, there are some previous works \cite{Bruna2014ICLA,Duvenaudy2015CNG,Niepert2016LCN,Defferrard2016CNN} that discuss how to train CNN on graphs. These works \cite{Bruna2014ICLA,Defferrard2016CNN} use a spectral method by computing bases of the graph Laplacian matrix, and perform the convolution in the spectral domain. \cite{Niepert2016LCN} use a PATCHY-SAN architecture which maps from a graph representation onto a vector space representation by extracting locally connected regions from graphs.

\textbf{Deep learning on 3D shapes.} 
{ Recent works introduce various methods of deep architectures for 3D shapes. By using volumetric representation \cite{Wu2015ADR,Qi2016VMV}, standard CNN 2d convolution and pooling methods can be easily extended to 3D convolution and pooling. However, it leads to some problems which cannot be completely solved, such as data sparsity and the computation cost of 3D convolution. \cite{Kalogerakis2017SSP} proposed multi-view image-based Fully Convolutional Networks (FCNs) and surface-based CRFs to do the 3D mesh labelling by using RGB-D data as input. However, it may need a lot of time to compute the surface-pixel reference, and more viewpoints to maximally cover the shape surface. These methods \cite{Yi2017SSC,Su2017PND} proposed different deep architectures which can be trained directly on point clouds to achieve points segmentation and other 3D points analysis. However, the point cloud structure sampled from the origin meshes may lack the intrinsic structure of the meshes. For better understanding the intrinsic structures, many works directly adapt neural networks to meshes. The first mesh CNN method GCNN\cite{Masci20153DRR} used convolution filters in local geodesic polar coordinates. ACNN\cite{Boscaini2016NIPS} used anisotropic heat kernels as filters. MoNet\cite{Monti2017CVPR} considered a more generalized CNN architectures by using mixture model. \cite{Maron2017CNN} using a global seamless parameterization for which the convolution operator is well defined but can only be used on sphere-like meshes. As \cite{Boscaini2015SGP,Litany2017ICCV} show that intrinsic CNN methods can achieve better result on shape correspondence and descriptor learning with less training data. More details of these techniques can be found in  \cite{Bronstein2017SPM}.}

\textbf{Supervised methods for segmentation and labelling.} When using the supervised method to train a collection of labelled 3D shapes, an advanced machine learning approach is used to complete the related training \cite{Kalogerakis2010LTM,Guo:2015:TML,Xie2014TSS,Wang2012ACS,Wang2013PAT}. For example, \cite{Kalogerakis2010LTM} used Conditional Random Fields (CRF) to model and learn the sample example, in order to realize the component segmentation and labelling of 3D mesh shapes. \cite{Wang2013PAT} first projected 3D shapes to 2D space, and the labelling results in 2D space were then projected back to 3D shapes for mesh labelling. 
\cite{Xie2014TSS} use Extreme Learning Machines (ELM), which can be used to achieve consistent segmentation for unknown shapes. \cite{Guo:2015:TML,Kalogerakis2017SSP} applied deep architectures to produce the mesh segmentation. 

\textbf{Unsupervised segmentation and labelling.} A lot of research methods can build a segmentation model \cite{Huang2011JSS,Sidi2011UCS,Hu2012,Meng2013UCT,Xu2010SSA} and achieve joint segmentation without any label information. There are predominantly two unsupervised learning methods: matching and clustering. Using the matching method, the matching relation between pair 3D shapes is obtained based on the similarity of a relative unit given by correlation calculation. The segmentation shape of matching shape is then established to realize the joint segmentation \cite{Kreavoy2007,Huang2011JSS} of 3D shapes. By contrast, clustering methods analyze all the 3D shapes in the model set and cluster the consistent correlation units of 3D shapes into the same class. A segmentation model is then obtained and applied to consistent segmentation \cite{Hu2012,Meng2013UCT,Xu2010SSA}. 
% End By ShuiPP

\section{Shape Fully Convolution Network}%3

\begin{figure}[htbp]
\centering
\begin{minipage}[t]{0.32\linewidth}
\centering
\includegraphics[width=0.92\linewidth]{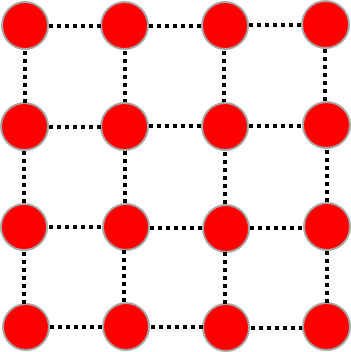}
\caption*{\emph{(a)}}%%add * to detete the Figure label
\end{minipage}
%\hspace{0.5ex}
\begin{minipage}[t]{0.32\linewidth}
\centering
\includegraphics[width=0.92\linewidth]{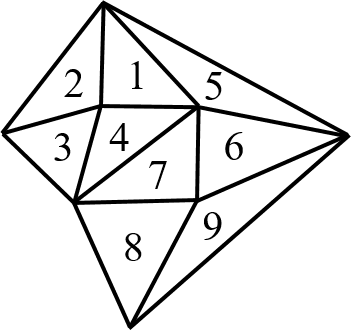}
\caption*{ \emph{(b)}}%%add * to detete the Figure label
\end{minipage}
%\hspace{0.5ex}
\begin{minipage}[t]{0.32\linewidth}
\centering
\includegraphics[width=0.92\linewidth]{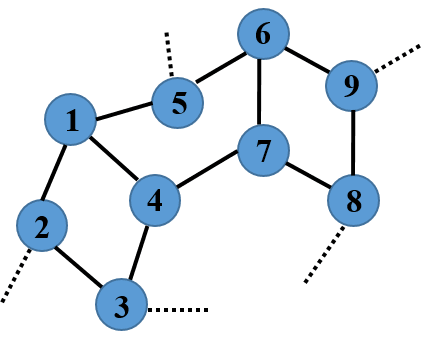}
\caption*{ \emph{(c)}}%%add * to detete the Figure label
\end{minipage}
\caption{\label{fig:fig2} Representation of different forms of data. (a) Image data representation; (b) Shape data representation; (c) Shape data represented as graph structure.}
\end{figure}

%Begin By ShuiPP
In the process of image semantic segmentation, it is mainly through operations such as convolution and pooling that fully convolution network architecture completes the image segmentation \cite{Long2015}. As analyzed above, the regular data structure amongst the pixels of the image makes it easy to implement these operations. By analogy with images, triangles of the 3D shape can be seen as pixels on the image, but unlike pixels, triangles of the shape have no ordered sequence rules. Figure \ref{fig:fig2}(a) represents the regular pixels on the image, while Figure \ref{fig:fig2}(b) represents the irregular triangles on the shape. It is difficult to complete the convolution and pooling operation on the 3D shape like that of the image. Therefore, based on the characteristics of 3D shape data structure and analogous to the operation on the image, we need to design a new shape convolution and pooling operation.

As a 3D shape is mainly composed of triangles and the connections among them, we can use graph structure to describe it. Each 3D shape can be represented as $G=(V,E)$, with vertex $v \in V$ for a triangle and edge $e \in E \subset V \times V$ for the connection of adjacent triangles. The small triangle shown in \ref{fig:fig2}(b) corresponds to the graph structure shown in Figure \ref{fig:fig2}(c). Based on the graph structure, we design and implement a \emph{shape convolution and pooling operation}, which will be detailed in the following section.

\subsection{Convolution on Shape}%3.1

Convolution is one of the two key operations in the FCN architecture, allowing for locally receptive features to be highlighted in the input image. When convolution is applied to images, a receptive field (a square grid) moves over each image with a particular step size. The receptive field reads the feature values of the pixels for each channel once, and a patch of values is created for each channel. Since the pixels of an image have an implicit arrangement, a spatial order, the receptive fields always move from left to right and top to bottom. Analogous to the convolution operation on the image, therefore, we need to focus on the following two key ideas when employing convolution operation on the shapes:
\begin{enumerate}
\setcounter{enumi}{0}
\item \emph{Determining the neighbourhood sets around each triangle for the convolution operation according to the size of the receptive field.}
\item \emph{Determining the order of execution of the convolution operation on each neighbourhood set.}
\end{enumerate}
For the first point, the convolution operation on the image is mainly based on the neighbourhood relationship between pixels. Accordingly, we need to construct locally connected neighbourhoods from the input shape. These neighbourhoods are generated efficiently and serve as the receptive fields of a convolutional architecture, permitting the architecture to learn shape representations effectively.

\begin{figure*}[htbp]
  \centering
  \includegraphics[width=0.85\linewidth]{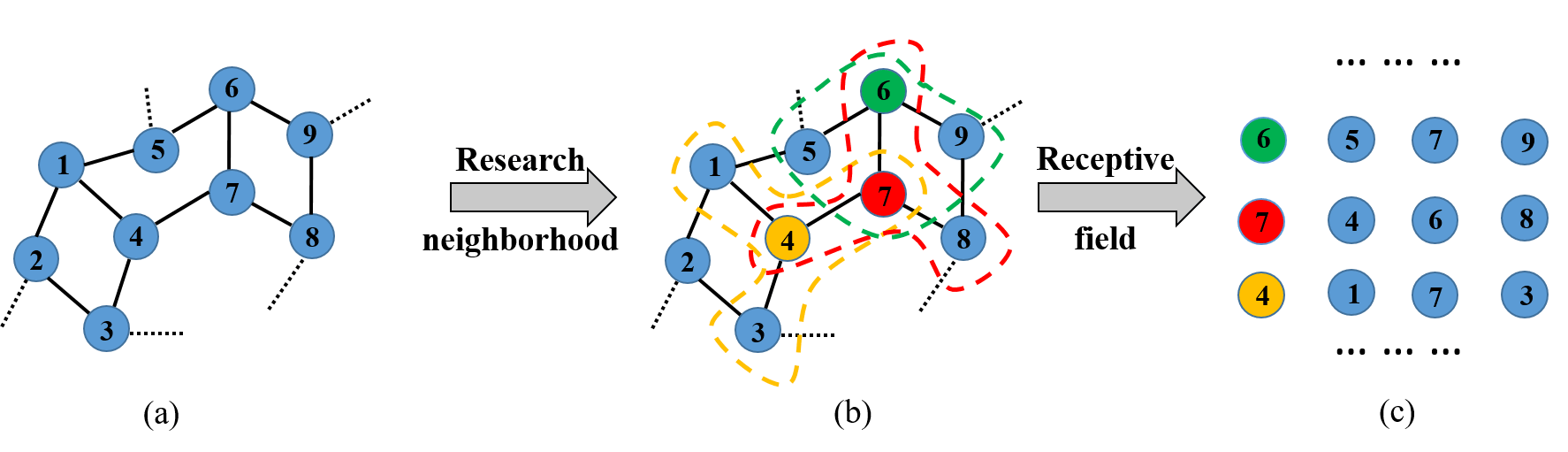}
  \caption{\label{fig:Fig3} Convolution process on a shape. (a) Shape represented as a graph; (b) The neighbourhood nodes of different source nodes searched by a breadth-first search, among which 4, 6, 7 represent source nodes. The areas circled by orange, red and green dotted lines are the neighbourhood nodes searched by each source node. (c) Convolution order of each neighbourhood set. The ellipsis is for other neighbourhood sets not represented.}
\end{figure*}
%\vspace{-5pt}

Shape has a neighbourhood relationship just like an image, but its irregularity restrains it to be directly represented and applied to the FCN learning. Yet when expressed as graph structures, the locally connected neighbourhoods of each triangle of the shape can be easily determined with various search strategies. In this paper, each triangle of the shape is viewed as a source node. We use the breadth-first search to expand its neighbourhood nodes on the constructed graph so as to obtain the neighbourhood sets of each triangle in the shape. Suppose the receptive field is set as $K$, the size of the neighbourhood set will be the same, including $K-1$ neighbourhood nodes and a source node, all of which will be used for a follow-up convolution operation. Figure \ref{fig:Fig3}(a) shows the graph structure of the shape, while Figure \ref{fig:Fig3}(b) shows the neighbourhood sets of each source node (that is, each triangle on the 3D shape) we obtained with the method outlined above.

As for the second point, when performing the convolution operation on the image, it is easy to determine the order of the convolution operation according to the spatial arrangement of the pixels. However, it is rather difficult to determine the spatial orders among triangles on the 3D shape. A new strategy is thus needed to reasonably sort the elements in the neighbourhood sets. Sorting is to ensure that the elements in each neighbourhood set can be convolved by the same rules, so that the convolution operation can better activate features.  For each node, all nodes in its neighbourhood set can be sorted by feature similarity (L2 similarity in the input feature space). Using this method, we can not only determine the order of the convolution operation of each neighbourhood set, but also ensure that the nodes in different sets have the same contribution regularity to their own source nodes in the convolution operation. The final convolution order for each neighbourhood set is shown in Figure \ref{fig:Fig3}(c). As shown in Figure \ref{fig:Fig3}(b), the execution order of the convolution operation of the neighbourhood set obtained from the source node is determined by calculating the input feature similarity.

\subsection{Pooling on Shape}%3.2

Pooling is the other key operation in the FCN architecture. The pooling operation is utilised to compress the resolution of each feature map (the result of the convolution operation) in the spatial dimensions, leaving the number of feature maps unchanged. Applying a pooling operation across a feature map enables the algorithm to handle a growing number of feature maps and generalises the feature maps by resolution reduction. Common pooling operations are those of taking the average and maximum of receptive fields over the input map \cite{Edwards2016GBC}. We share the same pooling operation on shape fully convolutional networks with the operation mentioned above. However, we need to address a key concern; that is, to \emph{determine the pooling operating order of the SFCN on the shape feature map.}

Similar to the convolution operation, we cannot directly determine the pooling operation order on SFCN based on spatial relationships among the triangles of the 3D shape. Since the 3D shape has been expressed as a graph structure, we can determine the pooling operation order according to the operation of convolutional neural networks on the graph. In this paper, the pooling operation on the SFCN is computed by adopting the fast pooling of graphs \cite{Niepert2016LCN,Defferrard2016CNN}.

\begin{figure*}[htbp]
  \centering
  \includegraphics[width=0.8\linewidth]{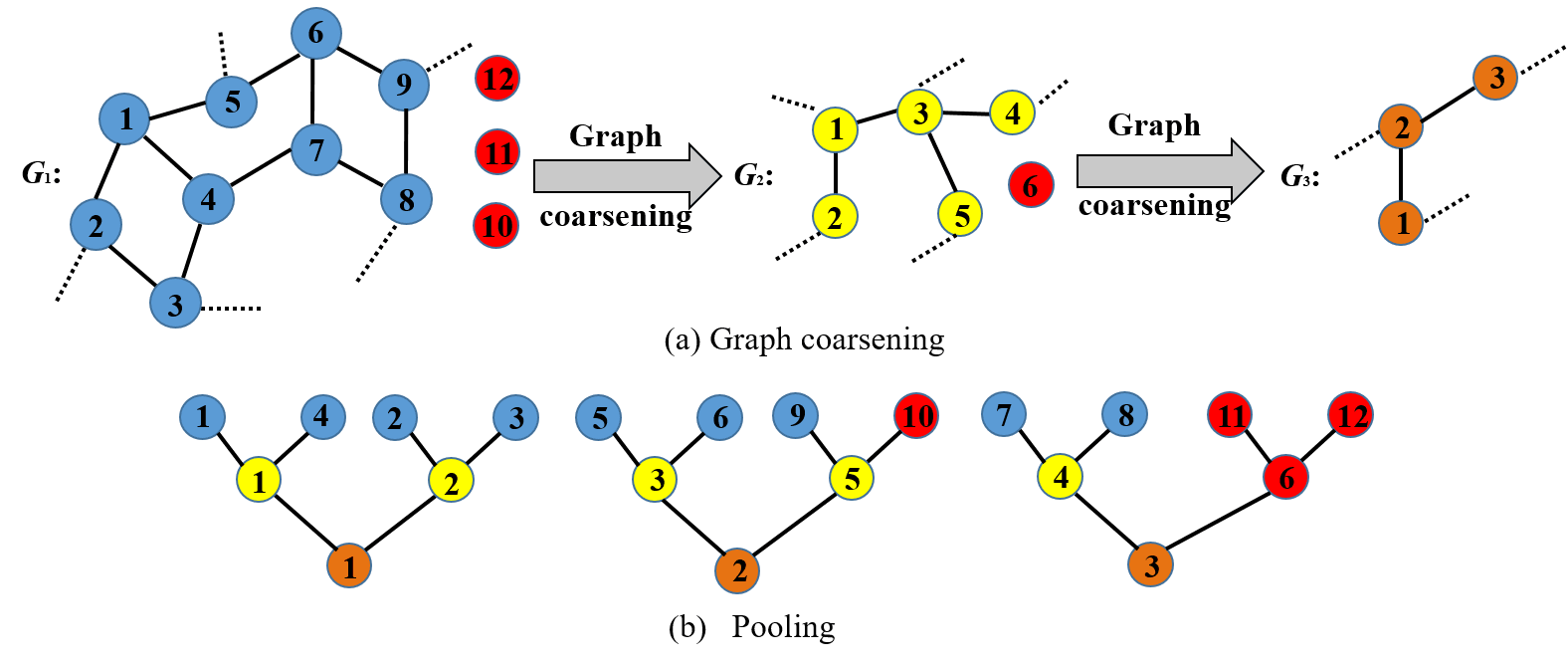}
  \caption{Example of graph coarsening and pooling. (a) Graph coarsening process. Note: the original graph has 9 arbitrarily ordered vertices. For a pooling of size 4, two coarsenings of size 2 are needed. To ensure that in the coarsening process the balanced binary tree can be constructed, we need to add appropriate fake nodes through calculation, which are identified in red. After coarsening, the node order on $G_3$ is still arbitrary, yet it can be manually set. Then backstepping the coarsening process, we can determine the node order of $G_2$ and $G_1$, and the corresponding relationship between nodes in each layer according to the order of $G_3$. At that point the arrangement of vertices in $G_1$ permits a regular 1D pooling. (b) Pooling order. The nodes in the first layer (blue and red) represent the $G_1$ node order; the second layer (yellow and red) represents $G_2$ node order; the third layer (purple) represents the $G_3$ node order and the corresponding relationship between nodes in each layer. The red nodes are fake nodes, which are set to 0 in the pooling process, as we carry out maximum pooling.}\label{fig:Fig4}
\end{figure*}

The pooling method for graphs \cite{Niepert2016LCN,Defferrard2016CNN} coarsens the graph with the Graclus multi-level clustering algorithm \cite{Dhillon2007WGC}. Graph coarsening aims to determine the new structure of the graph after pooling. We first present each shape as a graph structure, then we exploit the feature information on the triangles of the shape and the Graclus multi-level clustering algorithm \cite{Dhillon2007WGC} to complete shape coarsening; that is, to determine the new connection relationship of the shape feature map after pooling, which is shown in Figure \ref{fig:Fig4}(a). 

In the pooling process, traversing the feature map in a certain order according to the size of the receptive field is a key step to complete the calculation, namely, to determine the operation order of pooling. Following the method of pooling for the graph proposed by \cite{Defferrard2016CNN}, the vertices of the input graph and its coarsened versions are irregularly arranged after graph coarsening. To define the pooling order, therefore, a balanced binary tree is built by adding fake code to sort the vertices. Lastly, the pooling operation of the graph is completed based on the nodes order and the use of a 1D signal pooling method. After shape coarsening, we apply the same approach in this study to determine the order of pooling operations on the shape fully convolutional networks architecture, as shown in Figure \ref{fig:Fig4}(b). 

\section{Shape Segmentation via Shape Fully Convolution Network}%4

\vspace{-5pt}
% End By ShuiPP

%Begin By ShuiPP
We design a novel shape segmentation method based on SFCN, analogous to the basic process of image semantic segmentation on FCN \cite{Long2015}. Firstly, we extract three categories of commonly used geometric features of the shape as an input for SFCN training and learning. Secondly, based on the shape convolution and pooling operation proposed in Section 3 and the basic process of image semantic segmentation on FCN, we design a lattice structure suitable for 3D shape segmentation. Through training and learning of the network, we produce triangles-to-triangles labelling prediction. Finally, we introduce the optimisation process of shape segmentation.

\subsection{Geometric Feature Extraction}%4.1

Our approach is designed to complete the network training and learning based on some common and effective low-level features. In this paper, therefore, we extract three from the existing commonly used ones as the main features for the network training and learning. The three features include: average geodesic distance (AGD) \cite{Hilaga2001TMF}, shape context (SC) \cite{Belongie2002SMO} and spin image (SI) \cite{Johnson1999USI}. These features describe the characteristics of each triangle on a shape from multiple perspectives well. We also found in the experiment that these three features are complementary, which will be analyzed in detail in the experimental part.

\subsection{Shape Segmentation Networks Structure}%4.2

As the convolution and pooling operations on a shape are different from those on an image, the FCN architecture originally used in image segmentation cannot be directly applied on the 3D shape. We modify the original FCN architecture according to the convolution and pooling characteristics, so that it can be conducted in shape segmentation.

Our training network is made up of four parts: \emph{convolution, pooling, generating and deconvolution layers}, as shown in Figure \ref{fig:Fig5}. The convolution layer corresponds to a feature extractor that transforms the input shape to multidimensional feature representation. The convolution operation is completed by the method proposed in section 3.1. The pooling layer is used to reduce the feature vector of the convolution layer, and expand its receptive field to integrate feature points in the small neighbourhood into the new ones as output. The pooling operation is completed by the method proposed in section 3.2.

\begin{figure}[htbp]
  \centering
  \includegraphics[width=0.8\linewidth]{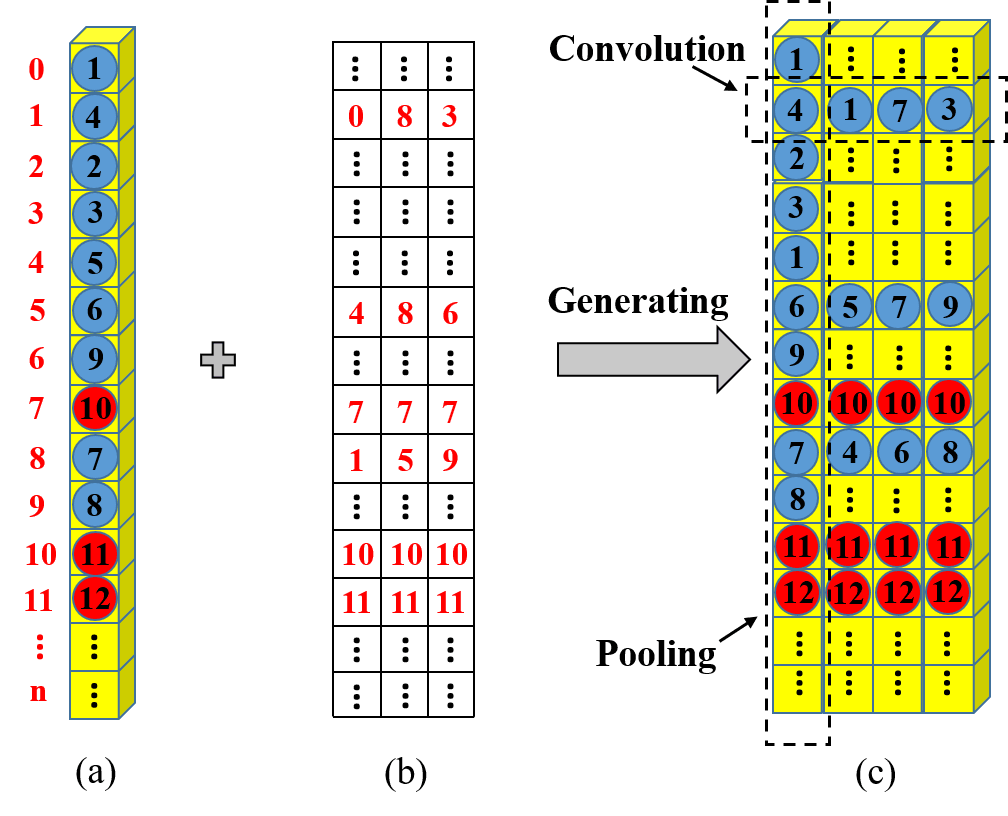}
  \caption{\label{fig:Fig6} Diagram of the generating layer. (a) The pooling order. A red number represents the offset, nodes in the blue circles represent nodes on the graph, nodes in the red circles are fake nodes. (b) Recorded neighbourhood set of each node. The numbers in the table represent the offset of each node after pooling sorting. (c) Data storage of the generating layer proposed in this paper, based on which our method can implement convolution operation by row and pooling operation by column.}
\end{figure}

\begin{figure*}[htbp]
  \centering
  \includegraphics[width=0.7\linewidth]{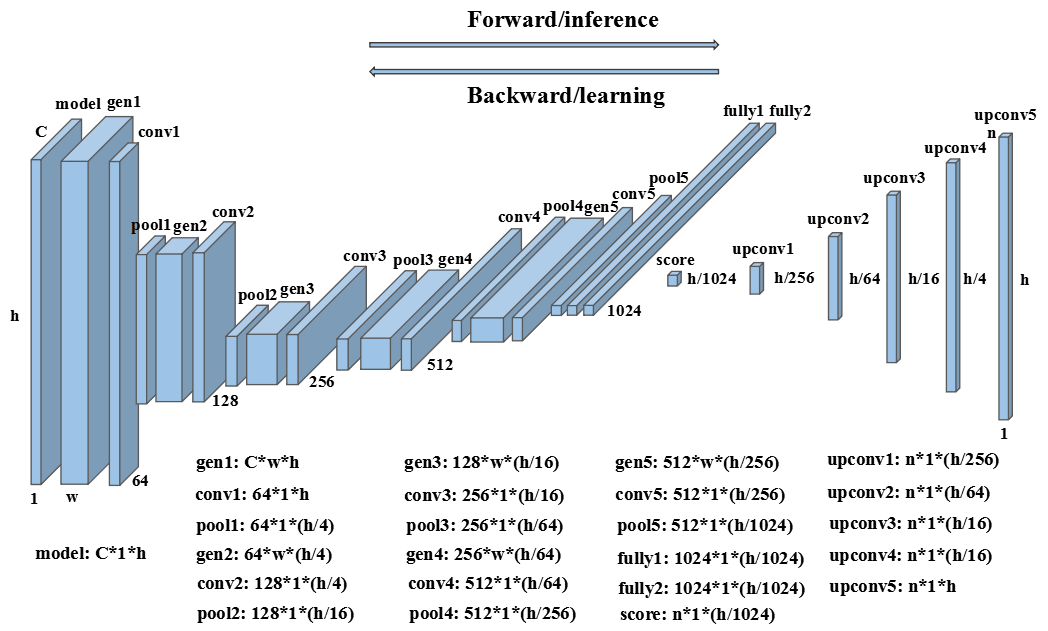}
  \caption{\label{fig:Fig5} The SFCN architecture we designed, mainly including 4 parts: convolution, pooling, generating, and deconvolution layers, which are represented by “gen”, “con”, “pool” and “deconv” respectively. Additionally, “fully” stands for fully connectivity and the numerical values reflect the dimension changes of each layer after calculation. C represents feature dimension, h represents the number of triangles of each shape and n represents the shape segmentation labels.}
\end{figure*}

As our convolution and pooling operations are designed for shapes, the original FCN architecture cannot be used directly. Compared with the original FCN's architecture, the architecture of the SFCN in this paper needs to record every neighbourhood set of each shape that participated in the convolution obtained in Section 3, as well as the convolution and pooling order of each shape. Thus, we add a \emph{generating layer} in the original FCN architecture, whose diagram of concrete meaning is shown in Figure \ref{fig:Fig6}. 

\vspace{-5pt}

Firstly, as shown in Figure \ref{fig:Fig6}(a), we can calculate the pooling order between nodes of the graph by the shape pooling method proposed in Section 3.2 (these nodes are equivalent to the triangle of the shape). We store the nodes in the calculation order on the \emph{generating layer}, as shown in Figure \ref{fig:Fig6}(c). Figure \ref{fig:Fig6}(a) gives the pooling order on a shape, where the figures represent the number of nodes (i.e. triangles). 

Secondly, we need to record the neighbourhood sets of each node (i.e. each triangle of the shape) involved in the convolution computation, as shown in Figure \ref{fig:Fig6}(b), where we store the neighbourhood sets by reading the offset in Figure \ref{fig:Fig6}(a). After the nodes are sorted by column on the \emph{generating layer}, we record their neighbourhood sets, in which the nodes are sorted according to the convolution order calculated in Section 3.1. As shown in Figure \ref{fig:Fig6}(c), each row that is sequenced in the convolution order represents a neighbourhood set of a node, where the figures represent the number of nodes (i.e. triangles). By storing the data in this way, we can achieve the convolution operation by row and the pooling operation by column, as shown in Figure \ref{fig:Fig6}(c). Another advantage of such storage is that, after pooling, the new neighbourhood set and the pooling order required by the next convolution layer of each new node can still be obtained and be applied to the next \emph{generating layer} using the method in Section 3.2. 

The deconvolution layer can be used to perform upsampling and densify our graph representation. As mention above, we regard pooling as a graph coarsening by clustering 4 vertices, as shown in Figure\ref{fig:Fig5}. Conversely, the upsampling operation can be regarded as a reverse process of the graph coarsening. We record how vertices are being matched before and after graph coarsening. Therefore it is easy to reuse the efficient deconvolution implementation based on the image-based FCN\cite{Long2015}. In this paper, the width of the convolution kernel in the deconvolution layer of the original FCN architecture is changed to 1 and the height is set to the size of the pooling we use, thereby obtaining the deconvolution layer of SFCN. The final output of the layer is a probability map, indicating the probability of each triangle on a shape that belongs to one of the predefined classes.

\begin{figure*}[htbp]
  \centering
  \includegraphics[width=0.8\linewidth]{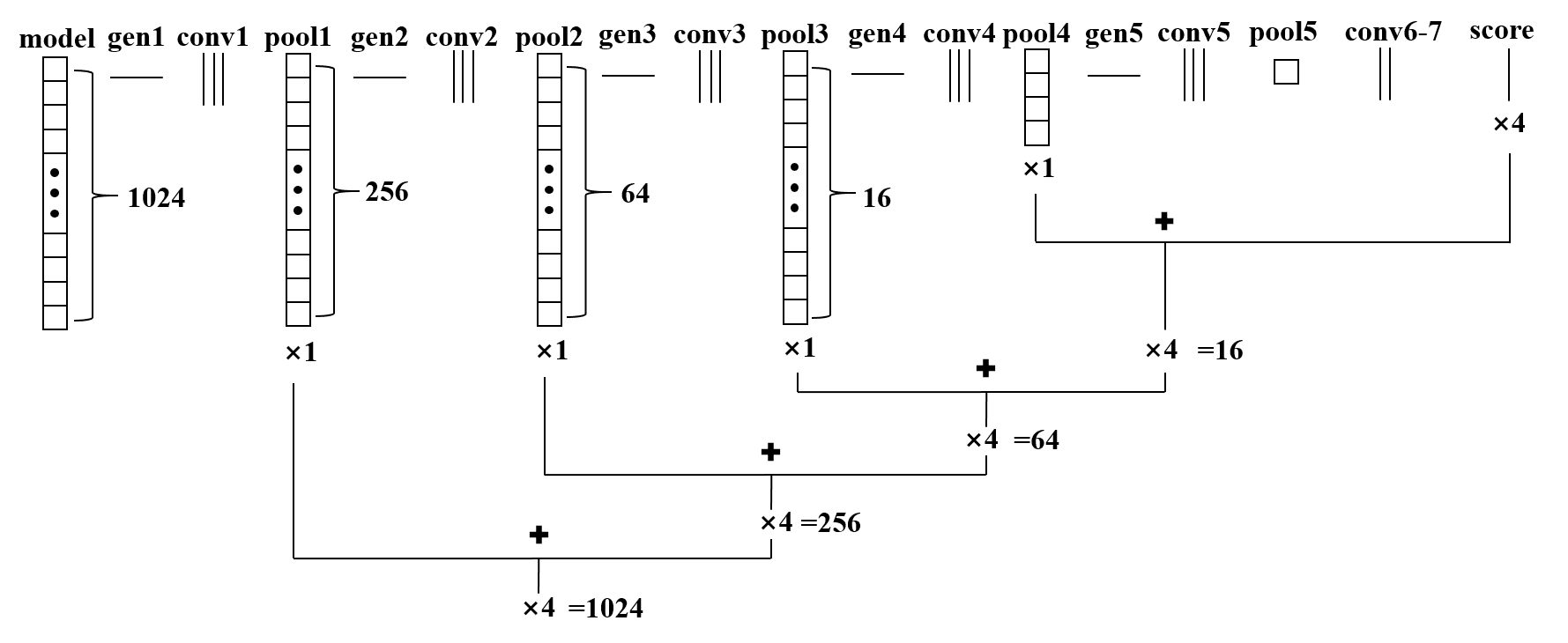}
  \caption{\label{fig:Fig7} The skip architecture for prediction. Our DAG nets learn to combine the coarse, high layer information with the fine, low layer one. We get the prediction results of the final layer by upsampling the score, then upsample again the prediction results combined with the final layer and the pool4 layer. After upsampling 5 times, we obtain the final prediction results, combining the final layer, pool5, pool4, pool3, pool2 and pool1 layer information, which then achieves triangles-to-triangles prediction.}
\end{figure*}

Based on the FCN architecture proposed by \cite{Long2015}, we design an SFCN architecture suitable for shape segmentation. Our convolution has five convolutional layers altogether, with each convolution layer having a generating layer before generating data for the next convolution and pooling, and followed by a pooling layer. Two fully connected layers are augmented at the end to impose class-specific projection. Corresponding to the pooling layer, there are five deconvolution layers, through which we can obtain the prediction probability of each triangle of the shape that belongs to each class. In the prediction process, we used the same skip architecture \cite{Long2015}. It can combine segmentation information from a deep, coarse layer with appearance information from a shallow, fine layer to produce accurate and detailed segmentations as the original FCN architecture. The specific process is shown in Figure \ref{fig:Fig7}. The prediction probability of each layer can be obtained by adding the results of a deconvolution layer and the corresponding results of the pooling layer after convolution, which also functions as the input for the next deconvolution layer. The number of rows will be repeated 5 times to return to the initial number of triangles, where the value of each channel is the probability of this class, realizing the triangle level prediction. Another difference from the original FCN architecture is that, in order to normalise the data, we add a batch normalisation layer after the convolution operation of the first layer of the original FCN architecture, using the default implementation of the BN in the Caffe \cite{J2014}.
\vspace{-5pt}

\subsection{Shape Segmentation Optimisation}%4.3

We train these three features separately using the network structure provided in Section 4.2. Given testing shapes, the SFCN produces the corresponding segmentation results under each feature, which can describe the segmentation of 3D shapes from different perspectives. Besides, due to the different input geometry features, there may be some differences among the predicted segmentation results of the same shape. To obtain the final segmentation results, we leverage the complementarity among features and the multi-label graph cuts method \cite{Boykov2001,Kolmogorov2004WEF,Boykov2004AEC} to optimise the segmentation results of the testing shape. The final segmentation result is obtained mainly through the optimisation of the following formula.

%set the space between equation and words
\setlength\abovedisplayskip{-1.2em}
\setlength\belowdisplayskip{-1.2em}
\begin{center}
\begin{equation}
%\begin{split}
\emph{E(l)} = \sum_{u \epsilon V} E_D(u,l_u) +  \sum_{\{u,v\} \epsilon E} E_S(u,v,l_u,l_v).
%\end{split}
\end{equation}
\end{center}
In this formula, $l_u$ and  $l_v$ are labels of triangles $u$ and $v$, data item $E_D(u,l_u)$ describes the energy consumption of triangle $u$ marked as label $l_u$, and smoothing item $E_S$ describes the energy consumption of neighboring triangles marked using different labels.

The first item of the formula is optimised mainly based on the probability that triangle $u$ is marked as label $l_u$. We predict the shape under the three features respectively, so the same triangle $u$ will have its own prediction probability under each feature. In this paper, utilising the feature's complementarity, we vote the labelling results to get the final prediction probability, and serve its negative logarithm similar to the paper \cite{Guo:2015:TML} as the first item of the multi-label graph cut. The second item in the formula smooths the segmentation results mainly through the calculation of the dihedral angle of the triangle and its neighbouring one. In this paper, the dihedral angle multiplied by the side length makes the second item of the formula to complete optimisation. Energy \emph{\textbf{E}} is minimized by employing Multi-label Graph Cuts Optimisation\cite{Boykov2001,Kolmogorov2004WEF,Boykov2004AEC}, through which we can obtain more accurate shape segmentation results.
% End By ShuiPP

\section{Experimental Results and Discussion}%5

%Begin By ShuiPP
\textbf{\emph{Data.}} In this paper, deep learning is used to segment the shape. Therefore to verify the effectiveness of this method, we first tested 3 existing large datasets from COSEG \cite{Wang2012ACS}, including the chair dataset of 400 shapes, the vase dataset of 300 shapes and the alien dataset of 200 shapes. The experiment of the mixed dataset was also carried out on the Princeton Segmentation Benchmark (PSB) \cite{Chen2009ABT} dataset. In addition, to confirm the robustness of the above method, we also selected 11 categories in the PSB\cite{Chen2009ABT} dataset for testing, each of them containing 20 shapes. For COSEG datasets, we chose the groundtruth as in the paper\cite{Wang2012ACS}. For PSB datasets, we chose the groundtruth as in the dissertation\cite{Kalogerakis2010LTM}.

\textbf{\emph{SFCN Parameters.}} We trained by SGD with momentum, using momentum 0.9 and weight decay of 1e-3. We chose ReLU as the activation function. Dropout used in the original classifier nets is included. The per-triangle, unnormalised softmax loss is a natural choice for segmenting shapes of any size, with which we trained our nets.

\textbf{\emph{Computation Time.}} We used two Intel(R) Xeon(R) CPU E5-2620 v3 @ 2.40GHz with 12 cores and NVIDIA GeForce TITAN X GPU. In large datasets, when we used 50 shapes for training and the triangles of each shape range from 1000 to 5000, the training would take approximately 120 minutes (including the time to compute the input descriptors), and the test and optimisation of a shape took about 30 seconds.

\textbf{\emph{Results.}} In the experiment, we used the classification accuracy proposed by Kalogerakis \cite{Kalogerakis2010LTM} and Sidi \cite{Sidi2011UCS} for the quantitative evaluation of our method.

Firstly, to compare with the methods of \cite{Guo:2015:TML} in the COSEG's large dataset, we randomly selected 50 shapes for training from the chair dataset and 20 shapes for training from the vase dateset. We compared our method with three shape segmentation methods \cite{Sidi2011UCS,Kim2013LPT,Guo:2015:TML}. The results are presented in Table \ref{tab:Table1}. It should be noted that the results and data obtained by other methods come from \cite{Guo:2015:TML}. The table shows that the results obtained by our method outperform the existing methods, thus proving that our method is effective.

\makegapedcells
\setcellgapes{1pt}%set table row and collumn length
\begin{table}[htbp]
\centering
\caption{Labelling Accuracy of Large Datasets.}
\label{tab:Table1}
%\smallskip
\begin{tabular}{|p{1.3cm}<{\centering}|p{1.2cm}<{\centering}|p{1.2cm}<{\centering}|p{1.2cm}<{\centering}|p{1.2cm}<{\centering}|}
%\begin{tabular}{|c|c|c|c|c|}
\hline
  & TOG [2011] & TOG [2013] & TOG [2015] & Ours \\
\hline
Chair & 80.20\% & 91.20\% & 92.52\% & 93.11\% \\
\hline
Vase & 69.90\% & 85.60\% & 88.54\% & 88.91\% \\
\hline
\end{tabular}
\footnotesize \newline\newline\leftline{ToG[2011] is short for Sidi et al. [TOG 2011], ToG[2013] is short for Kim}
\newline\leftline{et al.[TOG 2013],ToG[2015] is short for Guo et al.[2015].}
\end{table}%this is the Table \ref{tab:Table1}%table citation
\vspace{-5pt}

\begin{figure*}[htbp]
  \centering
  \includegraphics[width=1.0\linewidth]{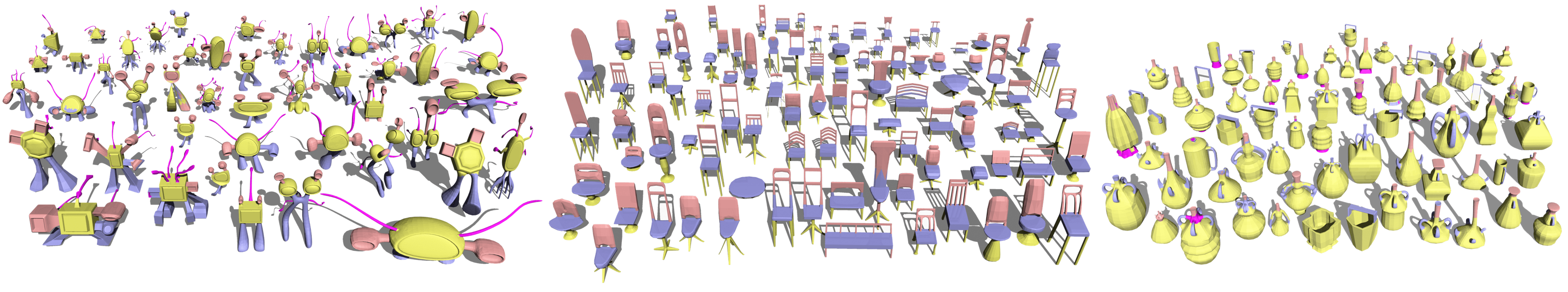}
  \caption{\label{fig:Fig8} Results under Large Datasets}
\end{figure*}
\vspace{5pt}

Secondly, to verify the effectiveness of our method for large datasets, we randomly selected 25\%, 50\% and 75\% of the shapes for training from each of the three large datasets from COSEG. To verify whether the SFCN prediction accuracy becomes higher as the number of training sets increases, we used the same 25\% shapes for testing in each experiment for each large dateset. In other words, the training sets gradually increased, while the test set remained the same. The results of this can be seen in Table \ref{tab:Table2}. Because the work of \cite{Xie2014TSS} carried out a similar experiment under a 75\% training condition with their method, we also make a comparison with their results in Table \ref{tab:Table2}. It can be seen from the table that the results obtained by our method perform better than theirs. Furthermore, with the increase of the training set, the classification accuracy of our method grows steadily in the same test set. This shows that with the increase of the training set, both the learning ability and the generalisation ability of the network architecture become stronger, which also proves the effectiveness of the designed network in this paper.

\makegapedcells
\setcellgapes{1pt}%set table row and collumn length
\begin{table}[htbp]
\centering
\caption{Labelling Accuracy of Large Datasets. Here are the results of the same test set for different numbers of training sets.}
\label{tab:Table2}
%\smallskip
\begin{tabular}{|p{1.5cm}<{\centering}|p{1.1cm}<{\centering}|p{1.1cm}<{\centering}|p{1.1cm}<{\centering}|p{1.4cm}<{\centering}|}
%\begin{tabular}{|c|c|c|c|c|}
\hline
  & Ours {}{} 25\% &  Ours {}{} 50\% & Ours {}{} 75\% & CGF[2014]  75\% \\ \hline
Chair& 93.43\% & 94.38\% & 95.91\% & 87.09\% \\ \hline
Vase & 88.04\% & 90.95\% & 91.17\% & 85.86\% \\ \hline
Tele-alien & 91.02\% & 92.76\% & 93.00\% & 83.23\% \\ \hline
\end{tabular}
\footnotesize \newline\newline\leftline{CGF[2014] is short for Xie et al.[CGF 2014]}
\end{table}%this is the Table \ref{tab:Table2}%table citation
\vspace{-5pt}

Thirdly, we visualize the segmentation results obtained by using our method in the three large datasets of COSEG, as shown in Figure \ref{fig:Fig8}. All the results shown are optimised and obtained using the 75\% training set. The segmentation results appear visually accurate, which proves the effectiveness of this method.

\begin{figure}[h]
  \centering
  \includegraphics[width=1.0\linewidth]{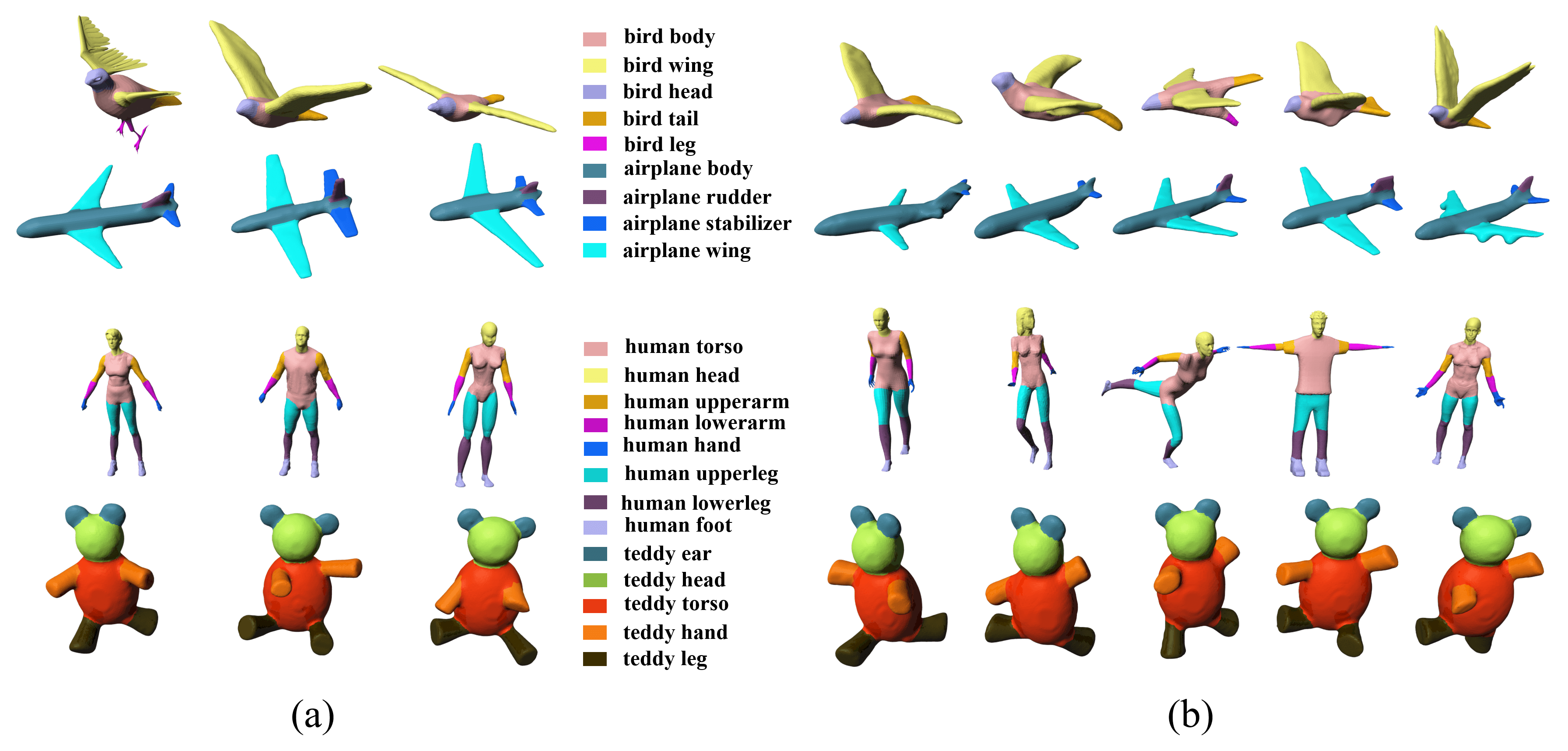}
  \caption{\label{fig:Fig9} The Segmentation Results of Mixed Datasets of Similar Shapes. (a) Part of the shapes in the training set; (b) Segmentation results of part of the shapes in the testing set.}
\end{figure}

Mixed dataset training and testing were performed as well. We respectively mixed airplane and bird, human and teddy, which are of similar classes. It must be noted that, unlike the method of \cite{Guo:2015:TML} which merges similar class labels, we retained these different labels. So the body of the plane and bird have different labels; their wings as well. 12 shapes of each dataset were selected as the training set, and the remaining as the test set. Our approach was repeated three times to compute the average performance, the segmentation results of which are shown in Figure \ref{fig:Fig9}. Figure \ref{fig:Fig9}(a) is part of the result of the training set while Figure \ref{fig:Fig9}(b) is part of the testing set. The classification accuracy of the two datasets is shown in Table \ref{tab:Table3}, which suggests that we can obtain good segmentation results when mixing similar data together. Although the segmentation accuracy is not as high as training, the average is above 90\%. Additionally, the basic segmentation is correct according to the visualisation of the results, proving that SPFCN network architecture is powerful in distinguishing features and learning.

\makegapedcells
\setcellgapes{1pt}%set table row and collumn length
\begin{table}[htbp]
\centering
\caption{Labelling Accuracy of Mixed Datasets of Similar Shapes}
\label{tab:Table3}
%\smallskip
\begin{tabular}{|p{2cm}<{\centering}|p{2cm}<{\centering}|p{2cm}<{\centering}|}
%\begin{tabular}{|c|c|c|}
\hline
  & Airplane \& Bird & Human \& Teddy \\ \hline
Accuracy & 90.04\% & 92.28\% \\ \hline
\end{tabular}
\end{table}%this is the Table \ref{tab:Table3}%table citation
\vspace{-5pt}

\begin{figure}[htbp]
  \centering
  \includegraphics[width=1.0\linewidth]{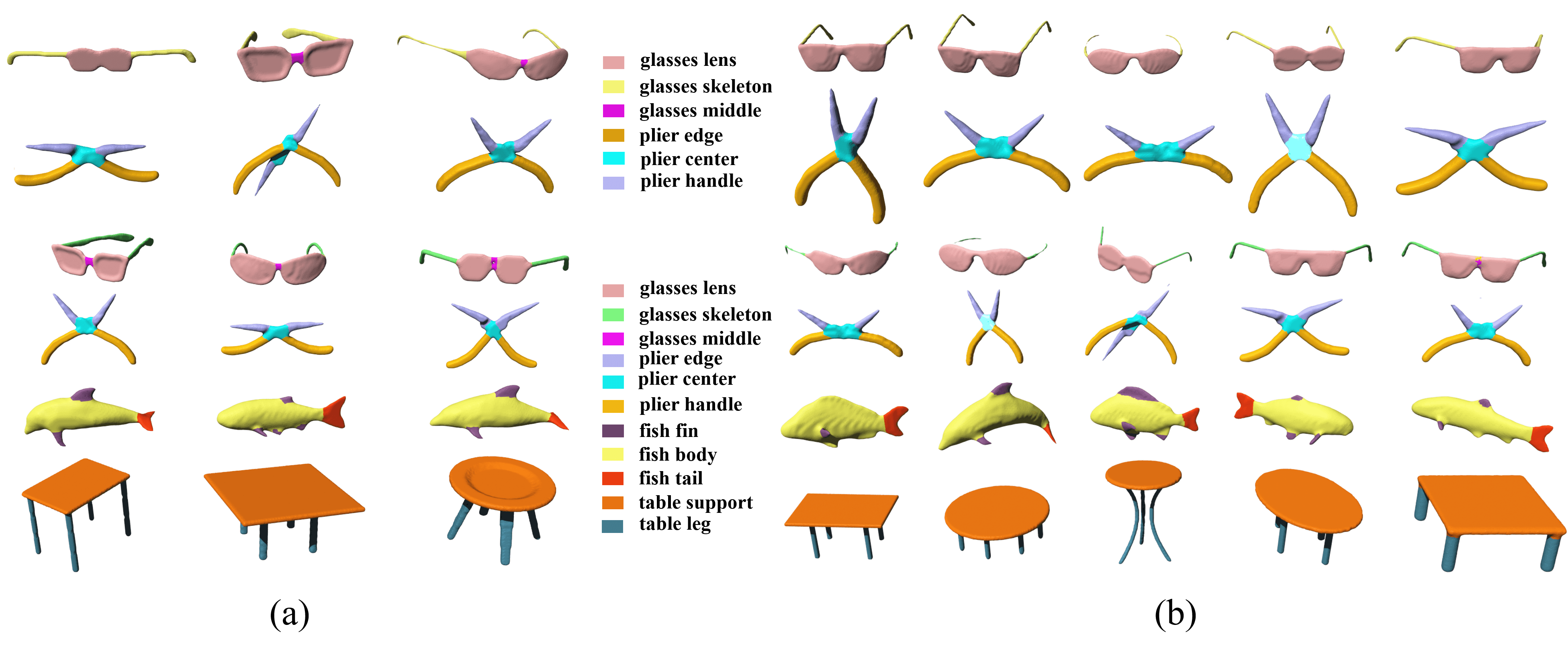}
  \caption{\label{fig:Fig10} The Segmentation Results of Mixed Datasets of Different Shapes. (a) Part of the shapes in the training set; (b) Segmentation results of part of the shapes in the testing set.}
\end{figure}

In addition, we mixed glasses and pliers, glasses, pliers, fish and tables of different categories. From the mixed datasets, we selected some data for training, with the remaining used for testing. Here 12 shapes of each dataset were selected as the training set, and the remaining as the test set. We also mixed two large datasets with 400 chairs and 300 vases. We selected 50\% of the shapes for training from each dataset and the remaining 50\% as the test set. Our approach was repeated three times to compute the average performance, the segmentation results of which are shown in Figure \ref{fig:Fig10} and Figure \ref{fig:Fig0}. Figure \ref{fig:Fig10}(a) is part of the result of the training set while Figure \ref{fig:Fig10}(b) is part of the testing set. The classification accuracy of the three datasets is shown in Table \ref{tab:Table4}, which also indicates that both the segmentation results and classification accuracy achieve impressive performance. In other words, this method can be used to train and predict any shape set, and proves once again that our SFCN architecture has good feature distinguishing ability and learning ability.

\makegapedcells
\setcellgapes{2pt}%set table row and collumn length
\begin{table}[htbp]
\centering
\caption{Labelling Accuracy of Mixed Datasets of Different Shapes}
\label{tab:Table4}
%\smallskip
\begin{tabular}{|p{1.2cm}<{\centering}|p{1.3cm}<{\centering}|p{1.7cm}<{\centering}|p{2.0cm}<{\centering}|}
%\begin{tabular}{|c|c|c|c|}
\hline
  & Glasses \& Plier & Chair(400)  \& Vase(300) & Glasses \& Plier \& Fish \& Table  \\ \hline
Accuracy & 96.53\% & 87.71\% & 91.82\% \\ \hline
\end{tabular}
\end{table}%this is the Table \ref{tab:Table4}%table citation
\vspace{-5pt}

\vspace{5pt}

\begin{figure}[htbp]
  \centering
  \includegraphics[width=1.0\linewidth]{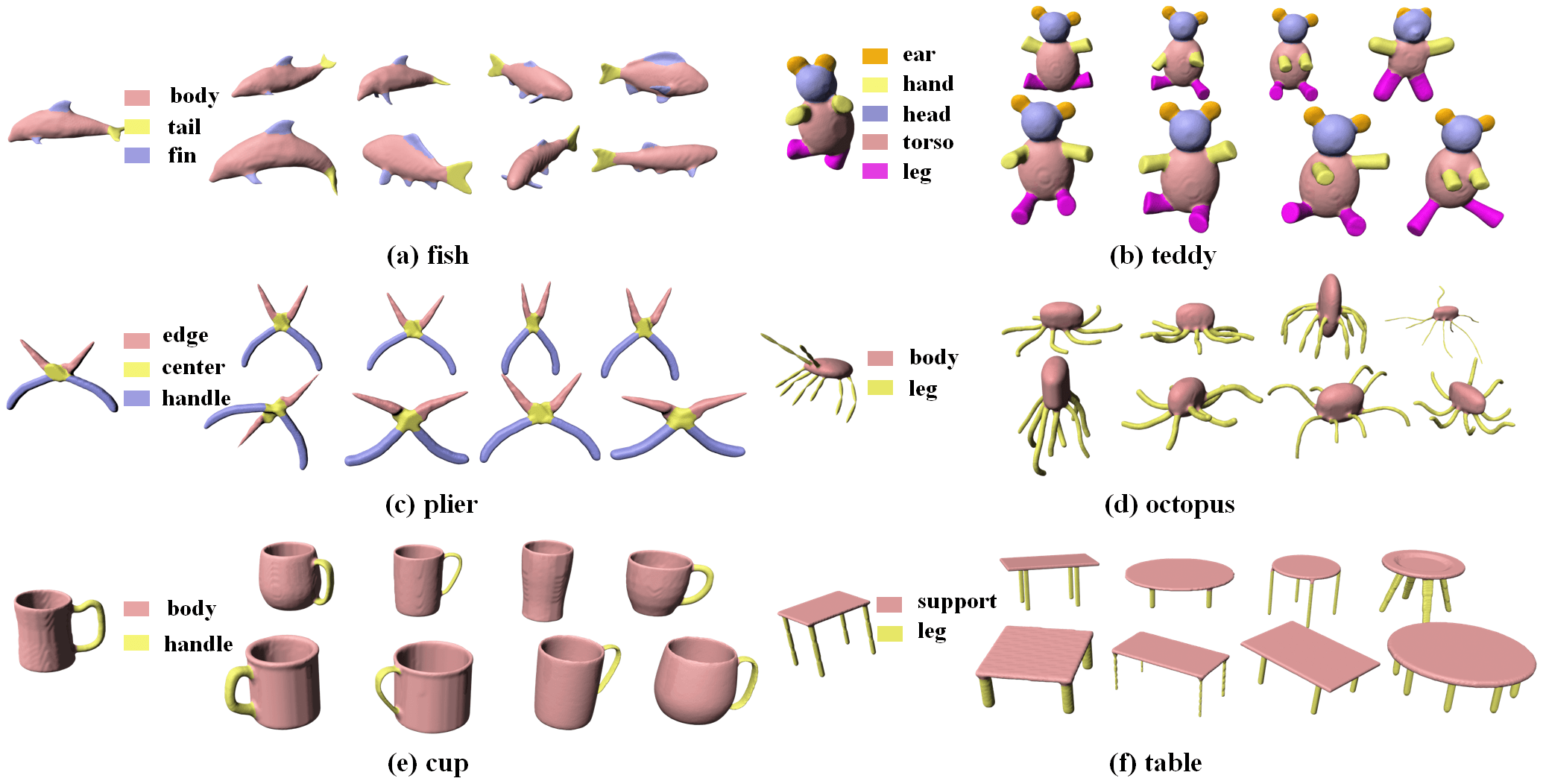}
  \caption{\label{fig:Fig11} More results of our method.}
\end{figure}

Lastly, as in the papers of \cite{Guo:2015:TML}, we separately trained several small datasets of PSB when $N=6$ and $N=12$ (i.e., SB6, SB12), in which $N$ is the number of randomly selected shapes in each training process. For each category of dataset, as in the papers of \cite{Guo:2015:TML}, we repeated our approach five times to compute the average performance. The comparison results with the related methods are presented in Table \ref{tab:Table5}. It should be noted that the results and data obtained by other methods come from \cite{Guo:2015:TML}. As shown in Table \ref{tab:Table5}, the results of several datasets of PSB obtained by our method perform much better in most categories than the existing methods \cite{Kalogerakis2010LTM,Guo:2015:TML,Xie2014TSS,Wang2013PAT,Chang2007LAL,Torralba2007SVF}, which proves the effectiveness of the method. On a few individual datasets, such as airplane, chair and table etc., our results do not go beyond those of other methods, yet are very close to the best ones, which also serves to prove that our method is effective. Moreover, the segmentation effect gradually strengthens as the training data increases, which shows that the learning ability of the SFCN architecture is enhanced with the increase of training samples. We also visualize the segmentation results of several datasets of PSB including teddy, pliers and fish etc. on the condition of SB12, the optimised ones of which are shown in Figure \ref{fig:Fig11}. Just like the large dataset, the results of the small one are visually accurate, which indicates that our method is feasible.

\vspace{-5pt}

\vspace{-5pt}

\makegapedcells
\begin{table*}[htbp]
\centering
\caption{Labelling Accuracy on Princeton Benchmark(SB6/SB12)}
\label{tab:Table5}
%\smallskip
\begin{tabular}{|p{1.0cm}<{\centering}|p{0.9cm}<{\centering}|p{1.3cm}<{\centering}|p{1.4cm}<{\centering}|p{1.4cm}<{\centering}|p{1.4cm}<{\centering}|p{1.0cm}<{\centering}|p{1.4cm}<{\centering}|p{1.4cm}<{\centering}|p{1.4cm}<{\centering}|p{1.0cm}<{\centering}|}
%\begin{tabular}{|c|c|c|c|c|c|c|c|c|c|c|}
\hline
%Class Name & SVM SB6 & JointBoost SB6 & ToG[2010]\footnotemark[1] SB6 & ToG[2013]\footnotemark[2] SB6 & ToG[2015]\footnotemark[3] SB6 & Ours & ToG[2010]\footnotemark[1] SB12 & ToG[2013]\footnotemark[2] SB12 & ToG[2015]\footnotemark[3] SB12 & Ours\\ \hline
  & SVM SB6 & JointBoost SB6 & ToG[2010]\tnote{1} SB6 & ToG[2013]\tnote{2} SB6 & ToG[2015]\tnote{3} SB6 & Ours SB6 & ToG[2010]\tnote{1} SB12 & ToG[2013]\tnote{2} SB12 & ToG[2015]\tnote{3} SB12 & Ours SB12\\ \hline
Cup & 94.11\% & 93.12\% & 99.1\% & 97.5\% & 99.49\% & \textbf{99.59\%} & 99.60\% & 99.60\% & 99.73\% & \textbf{99.74\%}\\ \hline
Glasses & 95.92\% & 93.59\% & 96.10\% & - & 96.78\% &  \textbf{97.15}\% & 97.20\% & - & 97.60\% & \textbf{97.79\%}\\ \hline
Airplane & 80.43\% & 91.16\% & 95.50\% & - & \textbf{95.56}\% & 93.90\% & 96.10\% & - & \textbf{96.67}\% & 95.30\%\\ \hline
Chair & 81.38\% & 95.67\% & 97.80\% & 97.90\% & \textbf{97.93}\% & 97.05\% & 98.40\% & \textbf{99.60}\% & 98.67\% & 98.26\%\\ \hline
Octopus & 97.04\% & 96.26\% & 98.60\% & - & 98.61\% &  \textbf{98.67\%} & 98.40\% & - & 98.79\% & \textbf{98.80}\%\\ \hline
Table & 90.16\% & 97.37\% & 99.10\% & \textbf{99.60}\% & 99.11\% & 99.25\% & 99.30\% & \textbf{99.60}\% & 99.55\% & 99.41\%\\ \hline
Teddy & 91.01\% & 85.68\% & 93.30\% & - & 98.00\% &  \textbf{98.04\%} & 98.10\% & - & 98.24\% & \textbf{98.27}\%\\ \hline
Plier & 92.04\% & 81.55\% & 94.30\% & - & 95.01\% &  \textbf{95.71\%} & 96.20\% & - & 96.22\% & \textbf{96.55}\%\\ \hline
Fish & 87.05\% & 90.76\% & 95.60\% & - & \textbf{96.22}\% & 95.63\% & 95.60\% & - & 95.64\% & \textbf{95.76\%}\\ \hline
Bird & 81.49\% & 81.80\% & 84.20\% & - & 87.51\% &  \textbf{89.03\%} & 87.90\% & - & 88.35\% & \textbf{89.48\%}\\ \hline
Mech & 81.87\% & 75.73\% & 88.90\% & 90.20\% & 90.56\% &  \textbf{91.72\%} & 90.50\% & 91.30\% & 95.60\% & \textbf{96.05}\%\\ \hline
Average & 88.41\% & 89.34\% & 94.77\% & 96.30\% & 95.89\% &  \textbf{95.98\%} & 96.12\% & \textbf{97.53}\% & 96.82\% & 96.85\%\\ \hline
\end{tabular}
\footnotesize \newline\newline\leftline{ToG[2010] is short for Kalogerakis et al. [2010], ToG[2013] is short for Wang et al. [2013],ToG[2015] is short for Guo et al.[2015].}
\end{table*}
\vspace{-5pt}

\begin{figure}[htbp]
  \centering
  \includegraphics[width=1.0\linewidth]{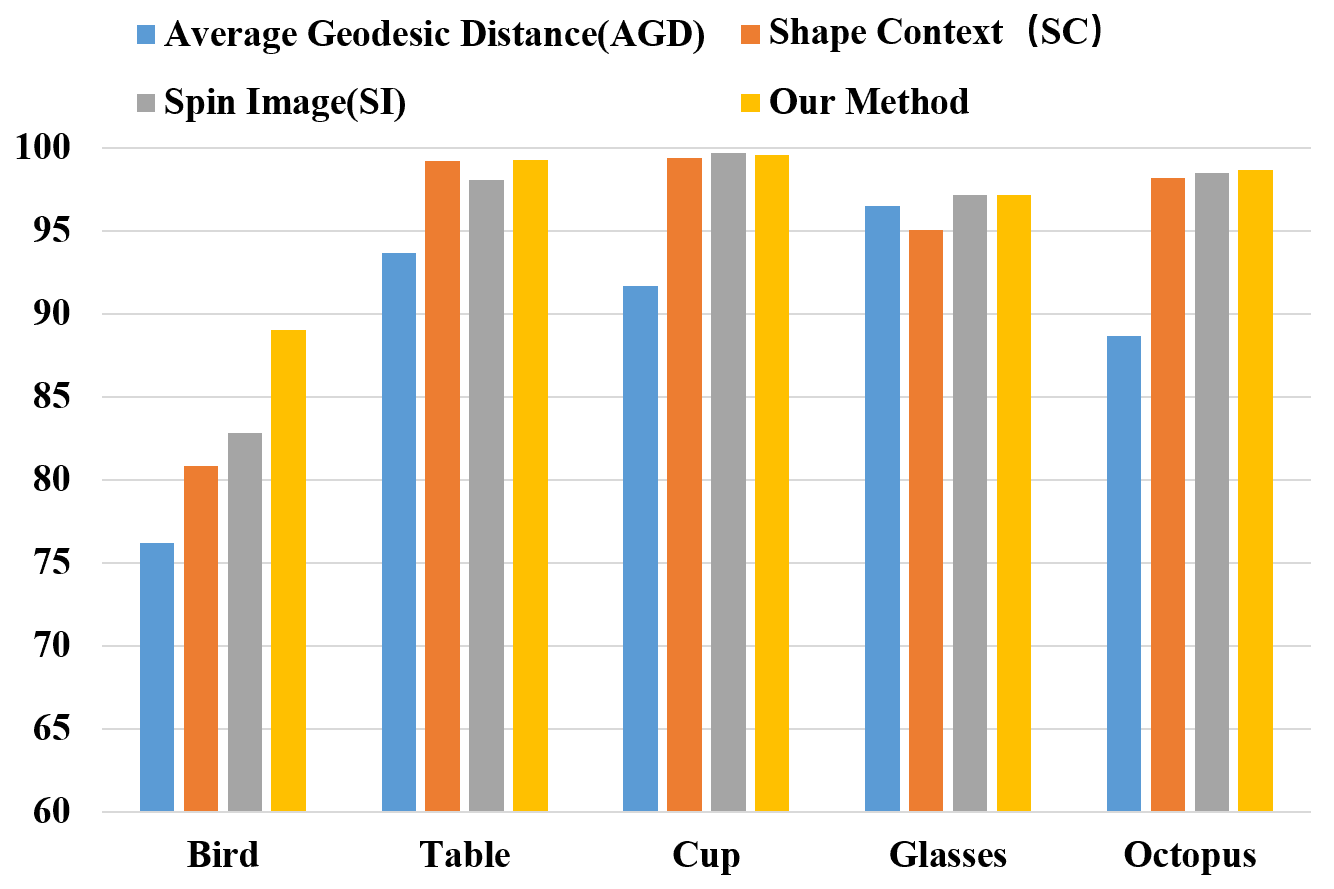}
  \caption{\label{fig:Fig12} The Comparison of Segmentation Accuracy under Different Features}
\end{figure}

\vspace{5pt}
\textbf{\emph{Feature sensibility.}} In this paper, we combine three features, average geodesic distance (AGD) \cite{Hilaga2001TMF}, shape context (SC) \cite{Belongie2002SMO} and spin image (SI) \cite{Johnson1999USI} to produce the shape segmentation. To verify the validity of this combination, we carried out a comparative experiment. The classification accuracy of each dataset of PSB under a single feature, and the one obtained with our method are compared in the condition of SB6, as shown in Figure \ref{fig:Fig12}. It can be seen that the classification accuracy of individual datasets under individual features is higher, but not significantly more so than that of our method. On the contrary, most datasets perform much better under our method. This shows that the features selected in this paper are complementary and the feature combination is effective.

\vspace{-5pt}

\begin{figure}[htbp]
  \centering
  \includegraphics[width=1.0\linewidth]{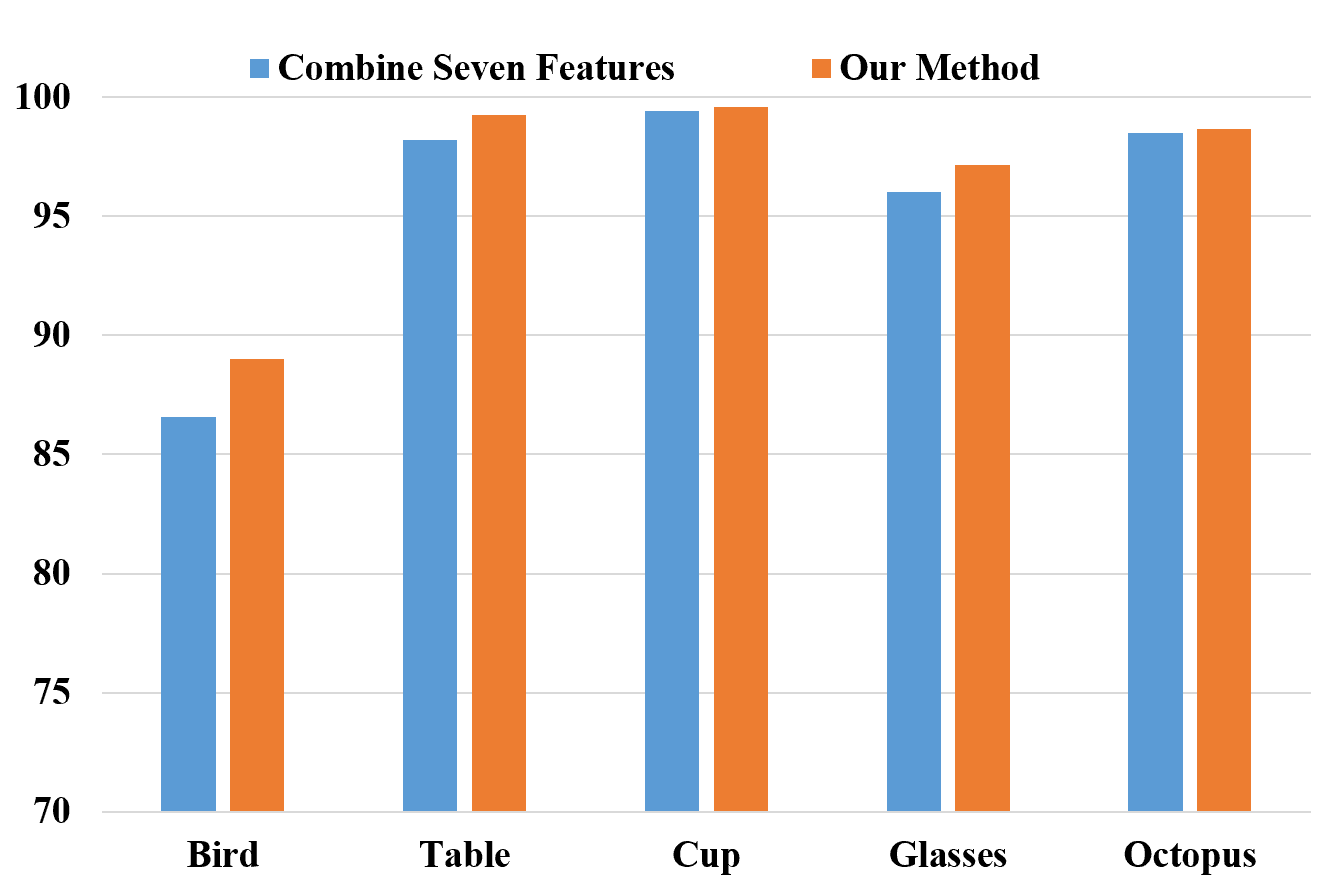}
  \caption{\label{fig:Fig13} The Comparison between Segmentation Accuracy under Three Features and under Seven Features}
\end{figure}

\vspace{5pt}
In the papers of \cite{Kalogerakis2010LTM} and \cite{Guo:2015:TML}, seven features are combined to perform segmentation experiments. In addition to the three features used in this paper, they also used four other features including curvature (CUR) \cite{Kavukcuoglu2009LIF}, PCA feature (PCA) \cite{Shapira2010CPA}, shape diameter function (SDF) \cite{Liu2009APS} and distance from medial surface (DIS) \cite{Long2015}. In this paper, several datasets of PSB are randomly selected to perform experiments with the combination of seven features. Under the same experimental conditions, the comparison results with the combination of three features are presented in Figure \ref{fig:Fig13}. The experiment tells us that for most datasets, the combination of three features brings higher classification accuracy than seven features, and for the remaining few datasets, the classification accuracy of the two is very close to each other. This indicates that the three features used in this paper can not only better complement, but also are more suitable for the network structure.  In addition, we also trained the combining three features in one SFCN network. However, it didn't produce a higher accuracy result and nearly doubled the training time. So, as a trade-off between performance and efficiency, we empirically trained 3 features separately to perform a segmentation task. 

\vspace{-5pt}

\begin{figure}[htbp]
  \centering
  \includegraphics[width=0.8\linewidth]{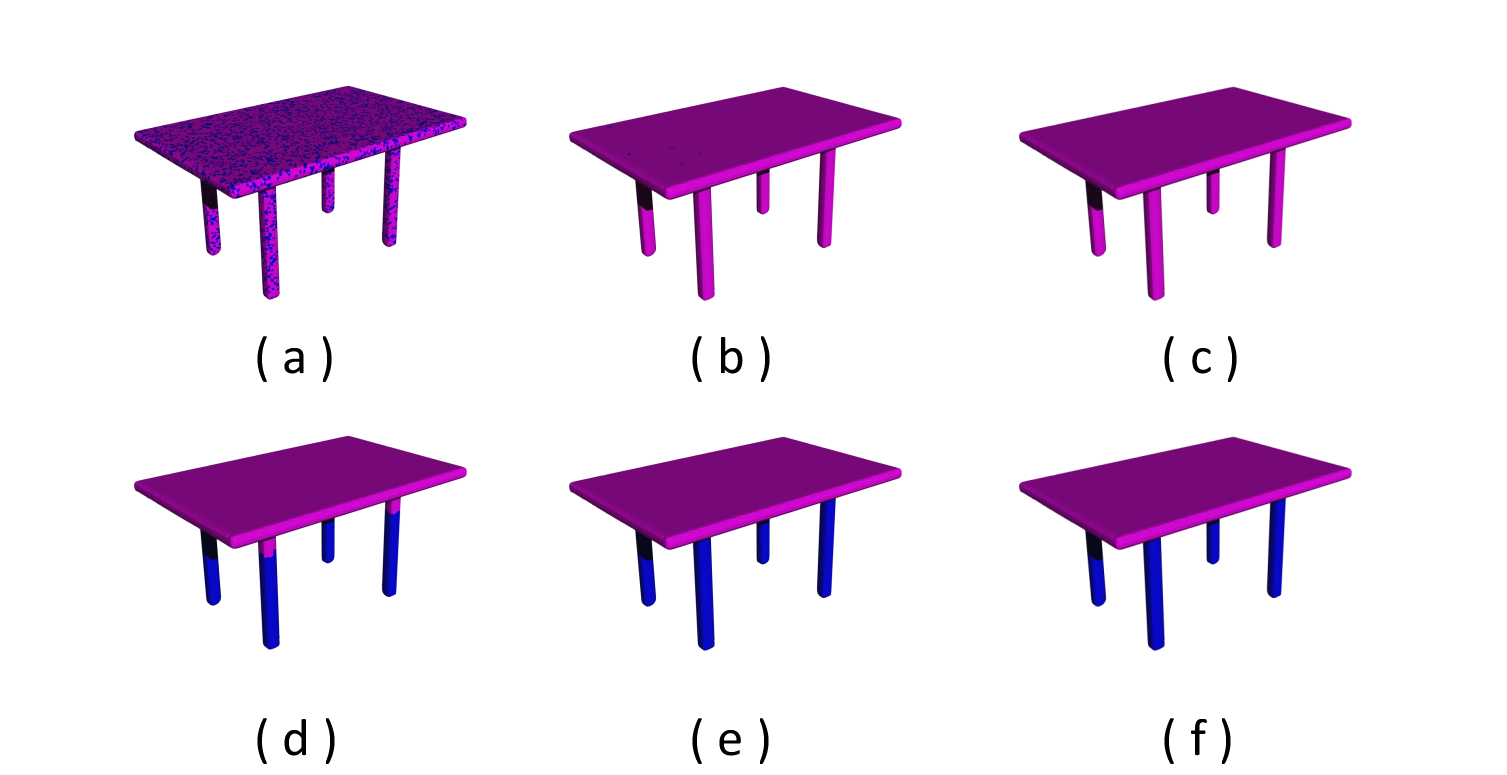}
  \caption{ The Segmentation Results of Layer Information with Different Steps. (a)- (e) Segmentation results Crossing 1 to 5 layers; (f) Groundtruth.}\label{fig:Fig16}
\end{figure}

\vspace{5pt}
\textbf{\emph{Skip architecture.}} During the SFCN training process, we utilised skip architecture of five layers and integrated the cross layer information with different steps to improve the SFCN segmentation results, and gradually improve the segmentation details. To verify the skip architecture can do mesh labelling learning and improve the segmentation results, we visualised the cross-layer prediction results with different steps. The segmentation results of several shapes crossing one to five layers are shown in Figure \ref{fig:Fig16}, where (f) is the groundtruth of the corresponding shape. Through tracking of the network computing process, we find that the network convergence is faster with the increase of the cross layers. In addition, it can be seen from the comparison results of cross layers and groundtruth in Figure \ref{fig:Fig16}, that with the increase of cross layers, the classification quality is gradually improved.

\begin{figure}[htbp]
  \centering
  \includegraphics[width=0.8\linewidth]{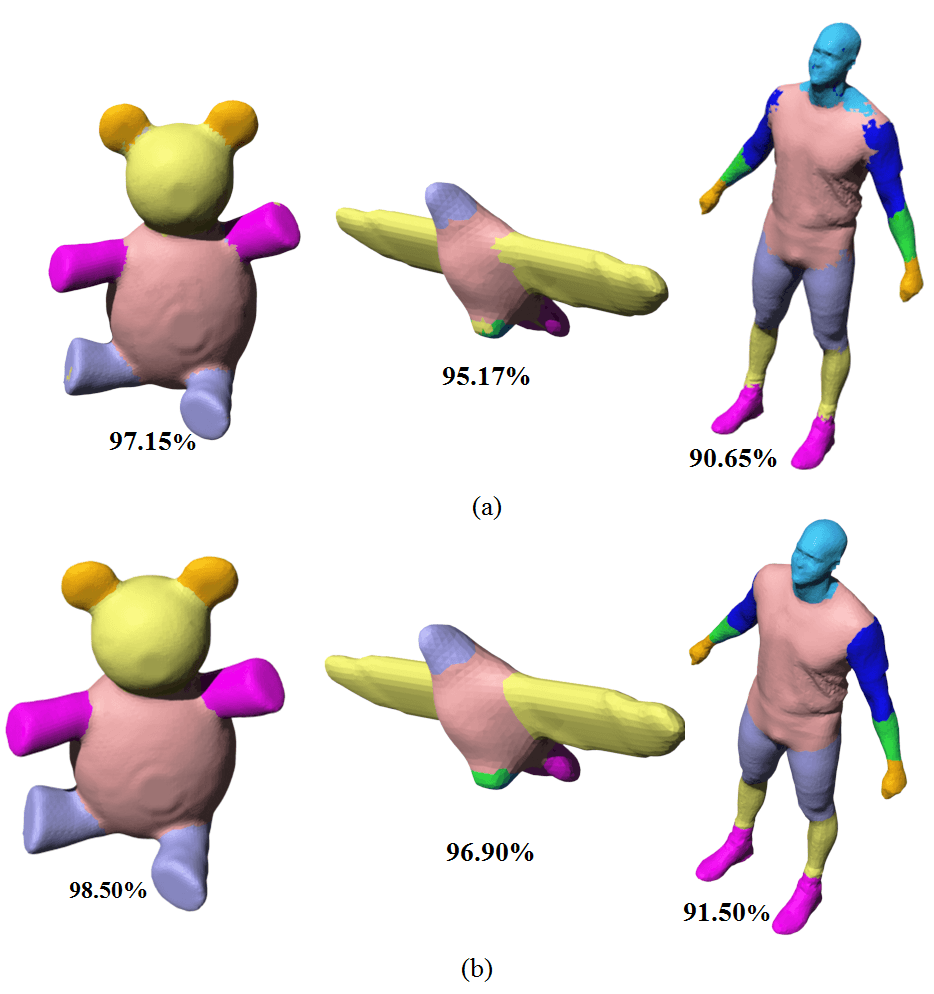}
  \caption{ Comparison Results before and after optimisation. (a) Segmentation results before optimisation; (b) Segmentation results after optimisation. The number below each shape is the segmentation accuracy.}\label{fig:Fig15}
\end{figure}

\vspace{5pt}
\textbf{\emph{Comparison before and after optimisation.}} In this paper, we use the multi-label graph cut optimisation method \cite{Boykov2001,Kolmogorov2004WEF,Boykov2004AEC} to optimise the segmentation results of testing shapes, based on the complementarity between features. The comparison results before and after optimisation of several shapes are shown in Figure \ref{fig:Fig15}. As shown in Figure \ref{fig:Fig15}(a), the results before optimisation are the final ones obtained by voting on the three different features tested by SFCN architecture. Figure \ref{fig:Fig15}(b) represents results after optimisation. As the results before optimisation are predicted in the triangle level, the boundary may be not smooth or there may be prediction errors of individual triangles in some areas. The above problems are well addressed after the optimisation with the multi-label graph cuts optimisation method \cite{Boykov2001,Kolmogorov2004WEF,Boykov2004AEC}, which proves that optimisation plays a significant role. In addition, the number below each figure is the classification accuracy of the corresponding shape, which shows that optimisation can improve classification accuracy.

\vspace{-5pt}

\vspace{-5pt}

\textbf{\emph{Limitations.}} Although the proposed method is effective at completing the 3D shape segmentation task, it has some limitations. Firstly, the shapes involved in the calculation must be manifold meshes, for they can easily determine the connection between triangles. Secondly, the designed SFCN architecture has no feature selection function, thus we carry on the separating training for the three features.{ Thirdly, to obtain better parameter sharing, the SFCN structure may need all training meshes to be the same triangulation granularity. Therefore, in future work, we will seek to improve the SFCN architecture, making it possible for automatic feature selection, and build an end-to-end structure.}

\section{Conclusions}%6
%Begin By ShuiPP
We design a novel shape fully convolutional network architecture, which can automatically carry out triangles-to-triangles learning and prediction, and complete the segmentation task with high quality. In the SFCN architecture proposed here, similar to convolution and pooling operation on images, we design a novel shape convolution and pooling operation with a 3D shape represented as a graph structure. Moreover, based on the original image segmentation FCN architecture \cite{Long2015}, we first design and implement a new generating operation, which functions as a bridge to facilitate the execution of shape convolution and pooling operations directly on 3D shapes. Additionally, accurate and detailed segmentation of 3D shapes is completed through skip architecture. To produce more accurate segmentation results, we optimise the segmentation results obtained by SFCN prediction by utilising the complementarity between the three geometric features and the multi-label graph cut method \cite{Boykov2001,Kolmogorov2004WEF,Boykov2004AEC}, which can also improve the local smoothness of triangle labels. The experiments show that the proposed method can not only obtain good segmentation results both in large datasets and small ones with the use of a small number of features, but also outperform some existing state-of-the-art shape segmentation methods. More importantly, our method can effectively learn and predict mixed shape datasets either of similar or of different characters, and achieve excellent segmentation results, which demonstrates that our method has strong generalisation ability.

In the future, we want to strengthen our method to overcome some limitations mentioned above and produce more results based on various challenging datasets, such as ShapeNet. Additionally, we hope that the SFCN architecture proposed can be applied to other shape fields, such as shape synthesis, line drawing extraction and so on.
%End By ShuiPP

\section*{Acknowledgements}
We would like to thank all anonymous reviewers for their constructive comments. This research has been supported by the National
Science Foundation of China (61321491, 61100110, 61272219) and the
Science and Technology Program of Jiangsu Province (BY2012190,
BY2013072-04).

%%Vancouver style references.
\bibliographystyle{cag-num-names}
\bibliography{refs}

\end{document}